\pdfoutput=1

\documentclass[final]{cvpr}

\usepackage{times}
\usepackage{epsfig}
\usepackage{graphicx}
\usepackage{amsmath}
\usepackage{amssymb}

\usepackage{float}

\usepackage{lipsum}
\usepackage{stfloats}
\usepackage{multicol}
\usepackage{multirow}
\usepackage{amsmath,bm}
\usepackage{etoolbox}

\usepackage{array}

\usepackage{tabulary}
\usepackage{xcolor}

\usepackage{paralist}

\usepackage{booktabs}

\definecolor{azure}{rgb}{0.0, 0.5, 1.0}
\definecolor{amethyst}{rgb}{0.6, 0.4, 0.8}
\newcommand{\BLUE}[1]{\textcolor{azure}{#1}}
\newcommand{\PURPLE}[1]{\textcolor{amethyst}{#1}}
\newcommand{\ORANGE}[1]{\textcolor{orange}{#1}}
\newcommand{\RED}[1]{\textcolor{red}{#1}}

\newcolumntype{?}{!{\vrule width 1pt}}

\usepackage[caption=false,font=footnotesize]{subfig}

\usepackage[pagebackref=true,breaklinks=true,colorlinks,bookmarks=false]{hyperref}
\hypersetup{colorlinks, citecolor=blue}
\hypersetup{colorlinks, linkcolor=blue}

\def\cvprPaperID{5425} 
\def\confYear{CVPR 2021}

\begin{document}

\title{Capturing Omni-Range Context for Omnidirectional Segmentation}

\author{Kailun Yang$^1$
~Jiaming Zhang$^1$
~Simon Reiß$^{1,2}$
~Xinxin Hu$^{3}$
~Rainer Stiefelhagen$^1$\\
\normalsize
$^1$CV:HCI, Karlsruhe Institute of Technology
\normalsize
~$^2$ZEISS
\normalsize
~$^3$Huawei Technologies\\
{\tt\small \{kailun.yang,jiaming.zhang,rainer.stiefelhagen\}@kit.edu}
}

\maketitle

\thispagestyle{empty}

\begin{abstract}
Convolutional Networks (ConvNets) excel at semantic segmentation and have become a vital component for perception in autonomous driving. Enabling an all-encompassing view of street-scenes, omnidirectional cameras present themselves as a perfect fit in such systems. Most segmentation models for parsing urban environments operate on common, narrow Field of View (FoV) images. Transferring these models from the domain they were designed for to 360$^\circ$ perception, their performance drops dramatically, \eg, by an absolute $30.0\%$ (mIoU) on established test-beds. To bridge the gap in terms of FoV and structural distribution between the imaging domains, we introduce Efficient Concurrent Attention Networks (ECANets), directly capturing the inherent long-range dependencies in omnidirectional imagery. In addition to the learned attention-based contextual priors that can stretch across 360$^\circ$ images, we upgrade model training by leveraging multi-source and omni-supervised learning, taking advantage of both: Densely labeled and unlabeled data originating from multiple datasets. To foster progress in panoramic image segmentation, we put forward and extensively evaluate models on \emph{Wild PAnoramic Semantic Segmentation (WildPASS)}, a dataset designed to capture diverse scenes from all around the globe. Our novel model, training regimen and multi-source prediction fusion elevate the performance (mIoU) to new state-of-the-art results on the public PASS ($60.2\%$) and the fresh WildPASS ($69.0\%$) benchmarks.~\footnote[1]{WildPASS: https://github.com/elnino9ykl/WildPASS}
\end{abstract}

\section{Introduction}
Convolutional Networks (ConvNets) reach striking performance on semantic segmentation~\cite{badrinarayanan2017segnet,long2015fully}, a dense visual recognition task that aims at transforming an image into its underlying semantic regions.
Particularly, segmenting images of road scenes automatically is of interest as it lies vital groundwork for scene understanding in an autonomous driving environment~\cite{jaus2021panoramic}.
Most segmentation algorithms~\cite{choi2020cars,orvsic2019defense,yang2018denseaspp} are designed to work on pinhole-camera images whose Field of View (FoV) is rather narrow, capturing only a fraction of what is occurring on and aside the road.
For a more holistic view on street-scenes, omnidirectional cameras are becoming ubiquitous in autonomous driving systems~\cite{yang2020omnisupervised}, as their capability of all-around sensing, gives rise to comprehensive 360$^{\circ}$ scene understanding.
\begin{figure}[t]
\setlength{\abovecaptionskip}{0pt}
\setlength{\belowcaptionskip}{0pt}
\centering
\includegraphics[width=0.48\textwidth]{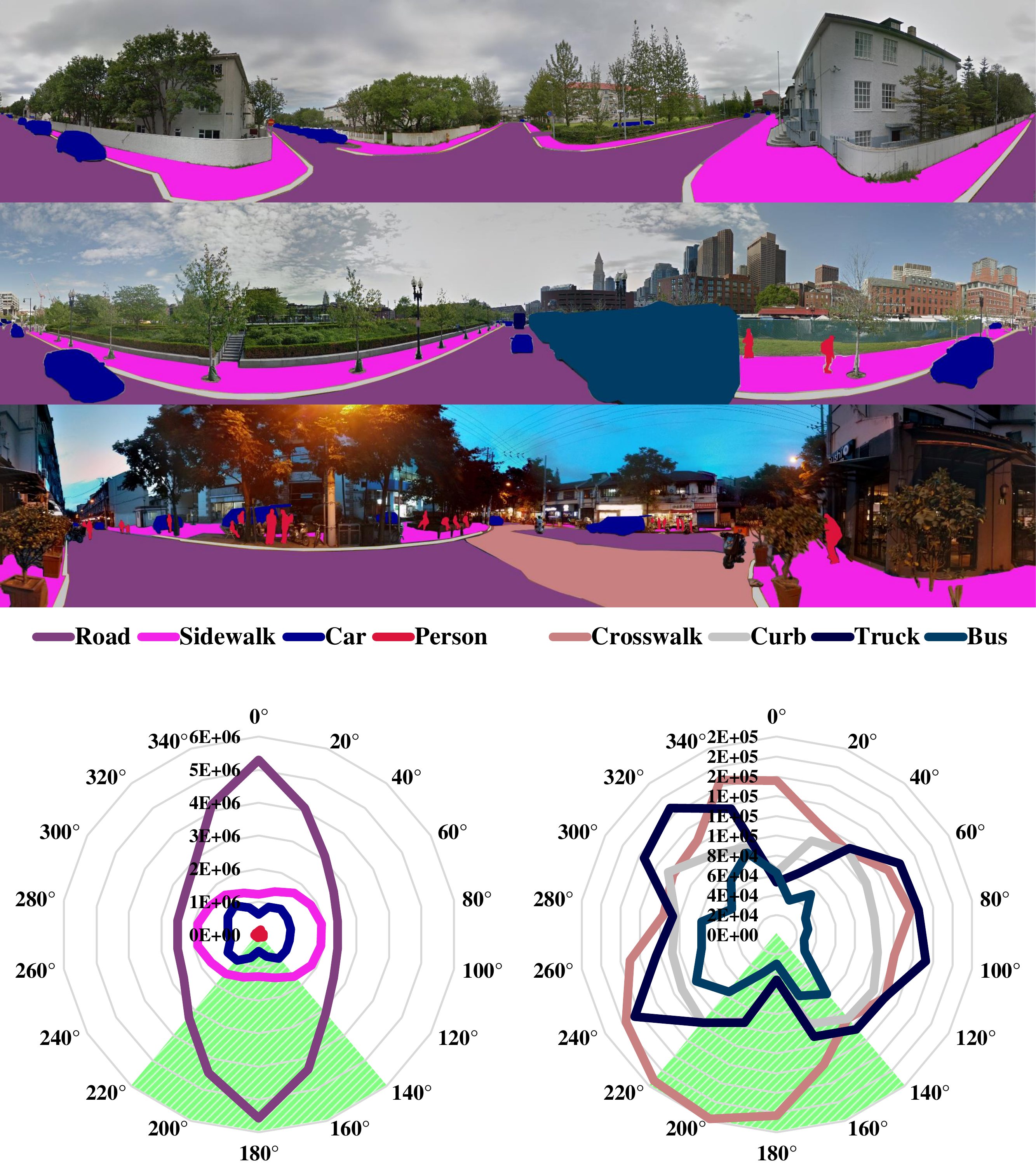}
\vskip-1ex
\caption{Top: Examples from our \emph{Wild PAnoramic Semantic Segmentation} (WildPASS) dataset; bottom: \emph{Class distribution} of pixel-class associations as unfolded over the angular direction.}
\label{figurelabel_concept}
\vskip-4ex
\end{figure}
Yet, rather than designing models based on the new imaging modality, predominantly, due to lack of sufficient labeled data, narrow FoV-trained models are applied to yield a segmentation~\cite{chen20193d,farabet2012learning,vasudevan2020semantic}.
Brought about by the large mismatch in FoV and structural distribution between imaging domains, this practice causes significant performance degradation, even to a point of rendering the perception of surroundings completely unreliable~\cite{yang2020ds}.

Fig.~\ref{figurelabel_concept} shows the distribution of semantic classes along the viewing angle, where green-tinted regions indicate the section visible to a front-facing narrow FoV camera.
It becomes apparent, that the positional priors in narrow- and omnidirectional images diverge severely, \eg, the road class occupies much of the center portion in narrow FoV images, which, in 360$^{\circ}$ images, is far smaller at viewing angles perpendicular to the front-facing direction (180$^{\circ}$).
These observations suggest distinct contextual priors between the imaging techniques, which provide critical cues for semantic segmentation~\cite{yu2020context,zhang2019co}.
Here, we address the challenging problem of leveraging inherent long-range contextual priors unique to omnidirectional images, that previous systems fail to harvest~\cite{yang2020pass}.
We lift the widely used, but computationally expensive non-local operations based on pixel-to-pixel correlations~\cite{fu2019dual,wang2018non} to the 360$^{\circ}$ FoV scenario in a computationally efficient fashion, \ie, modeling correlations of each position to omni-range regions.

By design, non-local attention~\cite{wang2018non} is meant to aggregate associations from highly correlated positions like regions of homogeneous classes, which often lie in the horizontal direction at a similar height, \eg, the sidewalks can be distributed across 360$^\circ$ but principally in the lower part of street-scenes. As preliminary investigation, we calculate the mean of the \emph{Pearson correlation coefficient} $r(X_{dir})$ of the probability distribution over the classes in either the horizontal $r(X_{hor})=0.37$ or vertical $r(X_{ver})=0.16$ direction.
In light of the substantially larger class-correlation of pixels observed horizontally, one can expect designing an attention module with emphasis in the large-FoV width-wise dimension to be advantageous, which reinforces the model's discriminability while eliminating redundant computation on weakly-correlated pixels belonging to heterogeneous semantics.
With this rationale, we propose \emph{Efficient Concurrent Attention Networks (ECANets)} for capturing omni-range dependencies within wide-FoV imagery.

Aside from considering the inherent properties of 360$^{\circ}$ data in architecture design, training such models is not straightforward, as a lack of sufficient annotated panoramic images impedes optimization.
The labeling process is extremely time- and labor-intensive, thus, only few surround-view datasets~\cite{sekkat202omniscape,wang2017torontocity,yogamani2019woodscape} are publicly accessible and none capture as diverse scenes as contemporary pinhole datasets~\cite{neuhold2017mapillary,varma2019idd} do.
This is why, to unlock the full potential of our novel ECANet architecture, we propose a \emph{multi-source omni-supervised learning regimen} that intertwines training on unlabeled, full-sized panoramic images and densely labeled pinhole images.
By means of data distillation~\cite{radosavovic2018data} and our novel multi-source prediction fusion, we increase prediction certainty, ingrain long-range contextual priors while still benefiting from large-scale, narrow FoV segmentation datasets.

To facilitate progress in omnidirectional semantic image segmentation, we put forward the \emph{Wild PAnoramic Semantic Segmentation} (WildPASS) dataset for evaluating models in unconstrained environments, featuring scenes collected from all around the globe, reflecting the perception challenge of autonomous driving in the real world.
An exhaustive number of diverse experiments demonstrate the proposed ECANet model, learning- and fusion strategy make it possible to deploy highly efficient ConvNets for panoramic image segmentation, surpassing the state-of-the-art on the public PASS~\cite{yang2020pass} and novel WildPASS datasets.
On a glance, we deliver the following contributions:
\begin{compactitem}
    \item Rethink omnidirectional semantic segmentation from the context-aware perspective and propose \emph{Efficient Concurrent Attention Networks}, capturing inherent long-range dependencies across the 360$^\circ$.
    \item Introduce \emph{multi-source omni-supervised learning}, integrating unlabeled panoramic images into training, empowering models to learn rich contextual priors.
    \item Present the diverse \emph{WildPASS dataset}: Collected from 6 continents and 65 cities, enabling evaluation of panoramic semantic segmentation in the wild.
    \item Our methods surpass previous and set new state-of-the-art results on the PASS and WildPASS datasets.
\end{compactitem}

\section{Related Work}
\noindent
\textbf{Context-aware semantic segmentation.}
Semantic segmentation has progressed exponentially
since the conception of fully convolutional networks~\cite{long2015fully}
and the groundwork laid by early encoder-decoder architectures~\cite{badrinarayanan2017segnet,noh2015learning,ronneberger2015u}.
Building atop classification networks~\cite{he2016deep,huang2017densely}, DeepLab~\cite{chen2017deeplab,chen2018encoder}, PSPNet~\cite{zhao2017pyramid}, RefineNet~\cite{lin2017refinenet} and DenseASPP~\cite{yang2018denseaspp} 
leverage pre-trained models and achieve notable performance improvements.
Yet, a lot of progress is driven by sub-modules like dilated convolutions~\cite{yu2015multi},
large kernels~\cite{peng2017large},
pyramid-, strip- and atrous spatial pyramid pooling~\cite{chen2017deeplab,hou2020strip,zhao2017pyramid}
for capturing structured context.

A second branch of work takes advantage of recent channel/spatial attention mechanisms~\cite{hu2018squeeze,hu2019acnet,wang2020eca,woo2018cbam} to exploit global context~\cite{li2018pyramid,zhang2018context}, dimension-wise priors~\cite{choi2020cars,yang2021context} and cross-modal features~\cite{sun2020real,xiang2021polarization,zhang2020issafe}.
More recently, inspired by non-local blocks in recognition tasks~\cite{wang2018non}, DANet~\cite{fu2019dual} and OCNet~\cite{yuan2018ocnet} adopt self-attention~\cite{vaswani2017attention} to capture either associations between any pair of pixels/channels or dense object context, respectively.
Further, point-wise attention is adaptively learned in~\cite{zhao2018psanet}. This trend prompts a variety of attention modules~\cite{chen20182,huang2020ordnet,li2019expectation,liu2020covariance},
as well as
graph-based models~\cite{yu2020representative,zhang2019dual}
and factorized variants~\cite{huang2019interlaced,sang2020pcanet,yin2020disentangled}.
With the nature of dense prediction in segmentation, the computational cost induced by combinatorics of pixel-pairs becomes infeasible very fast, which is amplified when working on pixel-rich, extra-wide panoramic images.
In order to reduce these pair-wise computations, CCNet~\cite{huang2019ccnet} operates in criss-cross paths, Axial-DeepLab~\cite{wang2020axial} propagates along height- and width-axis sequentially,
while~\cite{fu2020scene,shen2020ranet,yu2020representative,zhu2019asymmetric} act on key responding positions.
Addressing the nature of omni-range contextual priors in panoramic images, we propose an efficient concurrent attention module.
Unlike context-aware networks~\cite{hou2020strip,huang2019ccnet} that only aggregate spatial dependency or refine pixel-level context, our attention concurrently highlights horizontally-driven dependencies and collects global contextual information for wide-FoV segmentation. 
While accuracy-oriented networks reach high performance,
efficient networks like Fast-SCNN~\cite{poudel2019fast}, CGNet~\cite{wu2018cgnet} and ERFNet~\cite{romera2018erfnet} aim to perform both swift and accurate segmentation.
Further, a subset of compact ConvNets~\cite{wu2018cgnet,yang2020ds,yu2018bisenet} utilize attention mechanisms to efficiently aggregate context.
However, these methods do not consider inherent properties of omnidirectional imagery, and thus suffer from large accuracy degradation on such data.
Our solution to solve this issue extends the line of efficient ConvNets and consistently elevates their performance and reliability when operating on panoramic images.

\begin{figure*}[t]
\setlength{\abovecaptionskip}{0pt}
\setlength{\belowcaptionskip}{0pt}
\centering
\includegraphics[width=1.0\textwidth]{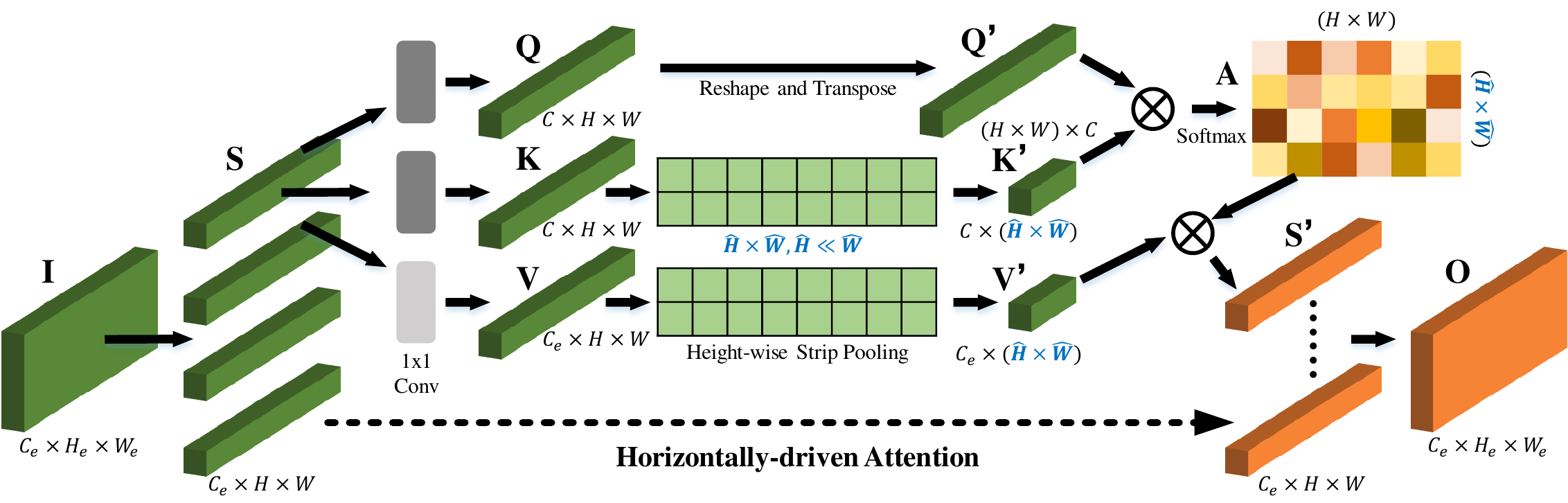}
\vskip-1ex
\caption{\emph{Horizontal Segment Attention (HSA)} module. The omnidirectional image feature is divided into segments to model self-attention for capturing horizontal contextual information. To further reduce computation complexities and emphasize dependencies along the 360$^\circ$ width-wise direction, the association is measured between each pixel and gathering regions obtained with height-wise strip pooling.}
\label{figurelabel_ha_attention}
\vskip-3ex
\end{figure*}
\noindent
\textbf{Omnidirectional semantic segmentation.}
As systems progressed towards 360$^{\circ}$ perception, early omnidirectional segmentation works were based on fisheye images~\cite{deng2017cnn,ye2020universal} or focused on indoor scenes~\cite{song2018im2pano3d,tateno2018distortion} with surround-view systems in street-scenes being largely based on a multitude of cameras~\cite{deng2019restricted,narioka2018understanding,pan2020cross,roddick2020predicting}.
Motivated by the prospect of attaining 360$^{\circ}$ semantic perception based on a singular panoramic camera, recent works build semantic segmentation systems directly on top of this sensor-modality~\cite{budvytis2019large,xu2019semantic,zhang2019orientation}.
However, these methods rely on either expensive manually labeled or synthesized omnidirectional data, that does not capture the diversity nor the realism as seen in large-scale pinhole image collections with rich ontologies~\cite{neuhold2017mapillary,varma2019idd}.
With this sentiment in mind, we introduce the WildPASS dataset, comprising 360$^{\circ}$ images from over 60 cities and multiple continents, encouraging a more realistic assessment of panoramic segmentation performance.
Yang \etal presented a framework~\cite{yang2020pass,yang2020ds} for re-using models trained on pinhole images via style transfer~\cite{zhu2017unpaired} and separating the panorama into multiple partitions for predictions, each resembling a narrow FoV image.
While quite accurate, running multiple forward passes suffers from high latency~\cite{yang2020pass} and disregards global context information within the panorama, as it is divided into patches.
With our holistic view on panoramic image segmentation, we cut this computational burden by nearly a factor of four, extending the framework in~\cite{yang2020omnisupervised} with a multi-source omni-supervised learning regimen that covers 360$^{\circ}$ imagery via data distillation~\cite{radosavovic2018data} as well as enjoying the advantages of large, readily available, labeled pinhole image datasets.
Further, making best use of the multi-source setting, we integrate a prediction fusion scheme, improving semantic certainty and generalization, as indicated by performance measures on both: PASS and WildPASS datasets.

\section{Methodology}
\subsection{Efficient Concurrent Attention}
\begin{figure}[t]
\setlength{\abovecaptionskip}{0pt}
\setlength{\belowcaptionskip}{0pt}
\centering
\includegraphics[width=0.48\textwidth]{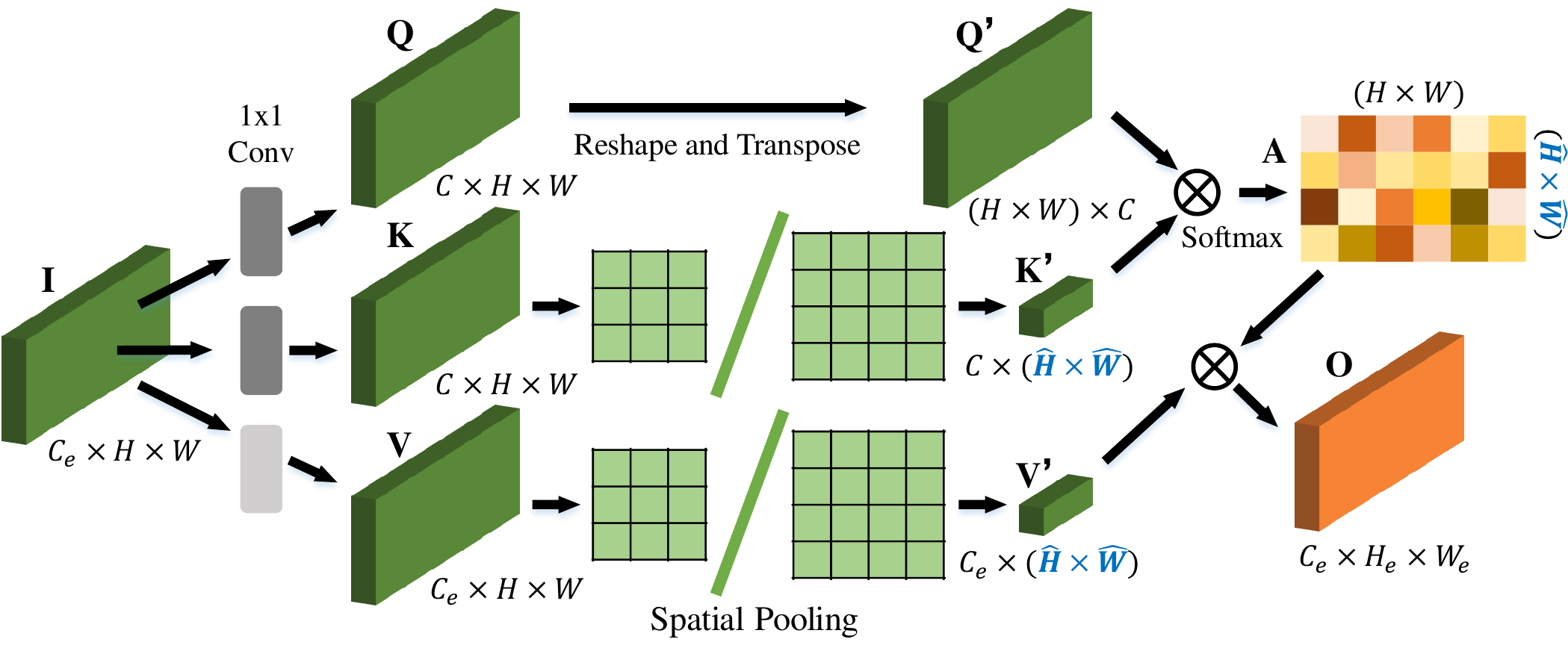}
\vskip-1ex
\caption{\emph{Pyramidal Space Attention (PSA)} module. To capture global context in an efficient way, self-attention is modeled with the association measured between each pixel and globally distributed regions sampled using spatial pooling of pyramid scales.}
\label{figurelabel_pa_attention}
\vskip-4ex
\end{figure}

\begin{figure*}[t]
\setlength{\abovecaptionskip}{0pt}
\setlength{\belowcaptionskip}{0pt}
\centering
\includegraphics[width=1.0\textwidth]{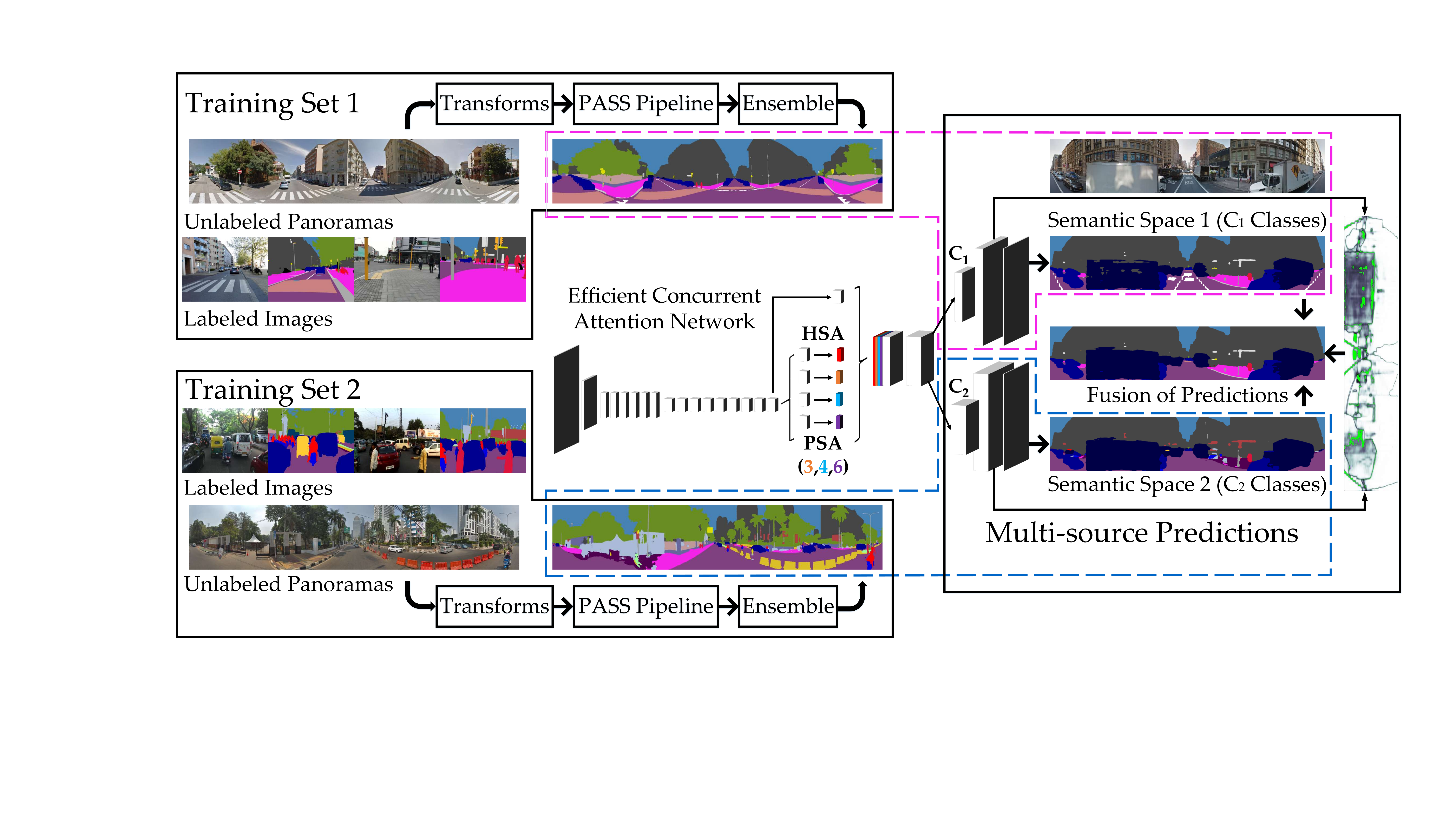}
\caption{Diagram of the \emph{multi-source omni-supervised learning regimen} for omnidirectional semantic segmentation. During training, the efficient concurrent attention network is trained on the union of labeled pinhole images and unlabeled panoramas in a multi-source manner. In the case of two domains, the network is enhanced with two heads for segmentation in heterogeneous label spaces. Annotations for the panoramas are automatically generated by seizing an ensemble of predictions on multiple transformations with the PASS pipeline~\cite{yang2020pass}. During testing, the efficient produces two sets of semantic maps with classes $C_1$ and $C_2$, which are fused in a post-processing step. On the right is an example of multi-space fusion, where green pixels on the uncertainty map denote areas with refined semantics (see the truck).}
\label{figurelabel_fusion}
\vskip-3ex
\end{figure*}

To capture omni-range contextual priors in panoramic images, we propose \emph{Efficient Concurrent Attention Networks (ECANets)}.
Classical non-local methods~\cite{fu2019dual,wang2018non} generate large affinity maps to measure the association among each pixel-pair, producing a complexity in space of $\mathcal{O}((H_e{\times}W_e)\times(H_e{\times}W_e))\sim\mathcal{O}({H_e}^2{W_e}^2)$.
$H_e{\times}W_e$ denotes the spatial resolution of the feature map, therefore at the rather small scale of 128$\times$64, the complexity already amounts to $67,108,864$ attention-related computations.
This substantially increases training footprint and inference requirement, thus hindering memory-efficient learning and fast execution.
With the assumption that wide-FoV panoramas contain rich contextual dependencies in the horizontal dimension, our ECANet features a \emph{Horizontal Segment Attention (HSA)} module (Fig.~\ref{figurelabel_ha_attention}) and a \emph{Pyramidal Space Attention (PSA)} module (Fig.~\ref{figurelabel_pa_attention}) for efficient context aggregation, simultaneously lessening the computational burden to a large extent as compared to the non-local baseline~\cite{wang2018non}.
As shown in Fig.~\ref{figurelabel_fusion}, the attended-to feature maps are passed through PSA at different scales and then concatenated with HSA- and backbone feature maps.
Finally, the resulting feature map is transformed by convolution- and upsampling layers to yield the semantic maps.

\noindent
\textbf{Horizontal segment attention module.}
The HSA module (Fig.~\ref{figurelabel_ha_attention}) produces horizontally-driven feature maps for a given input~${\rm\bf{I}}\in\mathbb{R}^{C_e{\times}H_e{\times}W_e}$ obtained from the backbone.
For street-scenes, wide-FoV panoramas contain rich dependencies in the spatial dimension~$W$, 
\eg, shared semantics between distant patches of sidewalk along the 360$^\circ$ can help in producing more consistent predictions.
In this sentiment and with the aim of modeling self-attention to capture the wide FoV dependencies, the input feature is first divided into $N$ segments ($N=4$ in Fig.~\ref{figurelabel_ha_attention}) along the $H$ dimension.
Each segment~${\rm\bf{S}}\in\mathbb{R}^{C_e{\times}H{\times}W}$ has a smaller size with $(H, W)=(\frac{H_e}{N},W_e)$, which is fed into convolution layers to generate feature maps of query $\rm\bf{Q}$, key $\rm\bf{K}$ and value $\rm\bf{V}$ to carry out self-attention~\cite{vaswani2017attention}.
This reinforces the horizontal context aggregation and reduces the complexity to $\mathcal{O}(N\times(\frac{H_e}{N}{\times}W_e)\times(\frac{H_e}{N}{\times}W_e))\sim\mathcal{O}({H_e}^2{W_e}^2/N)$.

Further, advocating 360$^\circ$ region-level priors and
to largely lessen computation burdens, we propose to apply a height-wise strip pooling on both the key feature~${\rm\bf{K}}\in\mathbb{R}^{C{\times}H{\times}W}$ and value feature~${\rm\bf{V}}\in\mathbb{R}^{C_e{\times}H{\times}W}$.
The resulting size of this strip pooling is $\widehat{H}\times\widehat{W}$, where $\widehat{H}\ll\widehat{W}$ but $\widehat{H}{\neq}1$ to enable interactions between vertically adjacent patches.
The attention map ${\rm\bf{A}}$, is then computed from the reshaped query feature ${\rm\bf{Q'}}\in\mathbb{R}^{(H{\times}W){\times}C}$ and the new key feature~${\rm\bf{K'}}\in\mathbb{R}^{C{\times}(\widehat{H}{\times}\widehat{W})}$:
\vskip-3ex
\begin{equation}
{\rm\bf{Q'}} \times
{\rm\bf{K'}} \rightarrow {\rm\bf{A}}\in\mathbb{R}^{(H{\times}W){\times}(\widehat{H}{\times}\widehat{W})}
\vspace{-1ex}
\end{equation}
In this manner, the association between each pixel and strip-pooled regions is ingrained in the module's architecture, which directly enables learning relations along the horizontal axis, across the 360$^\circ$.
We apply the softmax function to the attention map, transpose it and attend to the value feature~${\rm\bf{V'}}\in\mathbb{R}^{C_e{\times}(\widehat{H}{\times}\widehat{W})}$ as follows:
\vskip-2ex
\begin{equation}
{\rm\bf{V'}} \times
{\rm\bf{A^T}} \rightarrow
{\rm\bf{S'}}\in\mathbb{R}^{C_e{\times}(H{\times}W)}
\vspace{-1ex}
\end{equation}
Finally, all segments ${\rm\bf{S'}}\in\mathbb{R}^{C_e{\times}H{\times}W}$ are concatenated along the vertical dimension $H$ amounting to the horizontally-driven output~${\rm\bf{O}}\in\mathbb{R}^{C_e{\times}H_e{\times}W_e}$.
With this architectural design, we are able to greatly reduce the complexity to $\mathcal{O}(N{\times}(\frac{H_e}{N}{\times}W_e)\times(\widehat{H}\times\widehat{W}))\sim\mathcal{O}({H_e}{W_e}\widehat{H}\widehat{W})$, as $\widehat{H}\widehat{W}{\ll}{H_e}{W_e}$.
Further, our HSA module selectively aggregates context, as each pixel gathers information from its semantically-correlated regions in the horizontal direction.

\noindent
\textbf{Pyramidal space attention module.}
To capture global context, we introduce the PSA module (Fig.~\ref{figurelabel_pa_attention}), relating each pixel to globally distributed regions using spatial pooling. 
We depict the pyramid scales of the spatial pooling in Fig.~\ref{figurelabel_fusion} with $(\widehat{H},\widehat{W})=\{(3,3),(4,4),(6,6)\}$.
In this manner, the complexity is $\sum_{i}\mathcal{O}(H_eW_e\widehat{H}_i\widehat{W}_i)$, a considerable reduction as compared to the non-local baseline.
The attended-to feature maps are concatenated to the HSA module output and the features produced by the backbone.
Thereby, the proposed ECANet gathers both: wide-FoV horizontal semantics and global dependencies for enhanced omni-range reasoning.

\subsection{Multi-source Omni-supervised Learning}
In the following, we consider the training of panoramic segmentation networks and propose a better suited regimen.

\noindent
\textbf{Multi-source omni-supervision.}
We extend the design of~\cite{yang2020omnisupervised} based on data distillation~\cite{radosavovic2018data} and present a \emph{multi-source omni-supervised learning regimen}.
As shown in Fig.~\ref{figurelabel_fusion}, we train efficient networks on both labeled pinhole- and unlabeled panoramic images in a multi-source setup.
This adds diversity of FoV while simultaneously preventing the model from overfitting to pinhole images.
While different datasets have heterogeneous class definitions, therefore hindering direct training by merging sources, we argue that the relationships encoded in their similar label hierarchies positively affect the generalization of feature representations.
For example, street-scene datasets~\cite{cordts2016cityscapes,neuhold2017mapillary,varma2019idd} have different definitions of road surfaces, but share similar class hierarchies that both comprise flat region and traffic object categories.
As shown in Fig.~\ref{figurelabel_fusion}, our approach is designed to enable learning with unequal sources in spite of their class-label conflicts.
Precisely, to address the heterogeneity in semantic spaces, we append multiple output-heads to the efficient learner, each of which is a fully convolutional module with an upsampling layer for prediction in different label spaces (semantic spaces $1$ and $2$ for two sources, as illustrated in Fig.~\ref{figurelabel_fusion}).

For unlabeled panoramas, we create annotations using a sophisticated architecture pre-trained on pinhole sets by means of the PASS pipeline~\cite{yang2020pass}, as such we leverage the correspondence of FoV between pinhole images and partitioned panoramas for accurate segmentation.
The panorama is rotated along the horizontal direction based on its 360$^\circ$ wrap-around structure, and the predictions on differently rotated transformations are ensembled, producing the annotation.
While time-consuming due to the complex PASS pipeline and ensembling, the resulting fine-grained, high quality annotations are computed in advance and can be leveraged for data distillation (Fig.~\ref{figurelabel_fusion}).
Unlike~\cite{yang2020omnisupervised} that uses a battery of single-city unlabeled images, we distill knowledge via diverse panoramas from all around the globe, which will unfold as crucial detail for attaining generalization to open omnidirectional domains.

\noindent
\textbf{Multi-space fusion.}
In~\cite{yang2020omnisupervised}, while multi-space predictions are produced, only a single semantic map (\ie, the Vistas-space) is used for the inferred omnidirectional perception.
We assess this to be sub-optimal as it disregards the overlap in labels of the multi-source semantic spaces.
To address this, we propose to fuse the multi-space predictions in a post-processing step.
As such, we consider the intersecting classes of all semantic spaces~$\mathcal{C}_j$: $\mathcal{C}_{IN}=\bigcap_{j=1}^{J} \mathcal{C}_j$.
Based on the unnormalized output $\mathbf{z}_i^j$ for the $i^{\text{th}}$ pixel as produced by the trained $j^{\text{th}}$ output-head, the predicted class $y^j_i$ is known as~$ y_i^j=\operatorname*{arg\,max}_{c \in \mathcal{C}_j} \mathbf{z}_{i,c}^j$.
However, considering other candidate heads, the prediction will be updated as:
\vskip-2ex
\begin{equation}
    {y'}_i^j=\operatorname*{arg\,max}_{c \in \mathcal{C}_{IN}} \mathbf{z}_{i,c}^{j^*},
    \label{eq:prediction}
\vspace{-1ex}
\end{equation}
if the prediction of the optimal head $j^*$ is within the intersecting classes~$\mathcal{C}_{IN}$.
To find the optimal output-head $j^*$, several strategies can be leveraged.
A straightforward method based on the minimal uncertainty of the prediction is $j_{\text{var}}^* = \operatorname*{min}_{j\in J} (var(\mathbf{z}_i^{j}))$, where uncertainty is modeled by the variance $var(\mathbf{z}_i^{j})=(1/C_j){\sum_{c=1}^{C_j} (\mathbf{z}_{i,c}^{j} - \bar{\mathbf{z}}_{i}^{j})^{2}}$.
An exemplary uncertainty map is visualized on the right side of Fig.~\ref{figurelabel_fusion}.
Likewise, a common method for fusing predictions is to simply choose the maximum confidence based on the prediction probability: $j_{\text{conf}}^* = \operatorname*{max}_{j\in J}\max \sigma(\mathbf{z}_{i,c}^{j})$ with  $\sigma(\cdot)$ denoting the softmax function.
However, to select the optimal prediction, we make use of the calibrated confidence as given by the ratio of the highest and second highest probabilities: $j_{\text{cal}}^* = \operatorname*{max}_{j\in J} \frac{\max_{\mathbf{1}}\mathbf{z}_{i,c}^{j}}{\max_{\mathbf{2}} \mathbf{z}_{i,c}^{j}}$.
Compared to using all values in $j_{\text{var}}^*$ or just the maximum one in $j^*_{\text{conf}}$, this fusion refines panoramic segmentation by considering the top-2 predictions, as shown in Fig.~\ref{figurelabel_fusion}.

\section{Experiments}
\subsection{WildPASS Dataset}
\label{sec:wildpass}
\begin{figure*}[!t]
\setlength{\abovecaptionskip}{0pt}
\setlength{\belowcaptionskip}{0pt}
\centering
\includegraphics[width=1.0\textwidth]{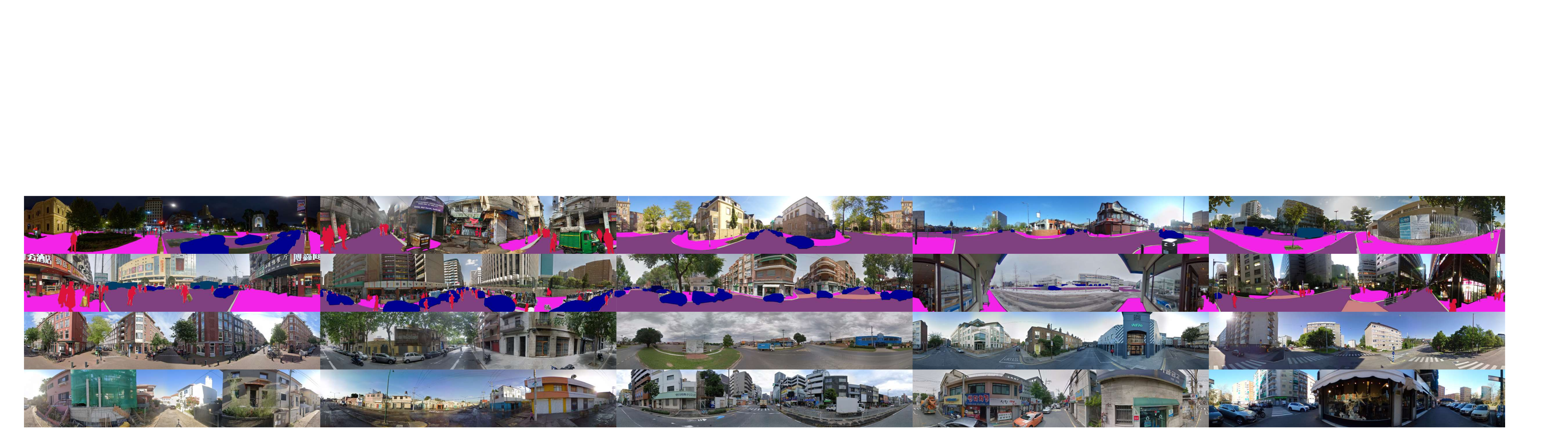}
\caption{Example images in the WildPASS dataset, including fine labeled and unlabeled panoramas collected from all around the world.}
\label{figurelabel_dataset}
\vskip-3ex
\end{figure*}
To foster research and
to facilitate credible numerical evaluation for panoramic scene segmentation, we introduce the \emph{Wild PAnoramic Semantic Segmentation} (WildPASS) dataset.
Unlike mainstream large-scale datasets like Cityscapes~\cite{cordts2016cityscapes} and BDD~\cite{yu2020bdd100k} that focus on urban scene understanding in Europe or North America, WildPASS embraces segmentation in the wild with images from all around the world based on Google Street View.
As shown in Fig.~\ref{figurelabel_dataset}, it includes unstructured and adverse road scenes like nighttime, as well as images
taken from different viewpoints to paint a comprehensive picture of real-world challenges for navigational perception systems.
Overall, we gather 2500 panoramas in 65 cities from all continents Asia, Europe, Africa, Oceania, North and South America (excluding only Antarctica).
Following PASS~\cite{yang2020pass} and RainyNight~\cite{di2020rainy}, 500 panoramas from 25 cities are precisely annotated for the most critical navigation-related classes.
We extend the semantics defined for panoramic segmentation~\cite{yang2020pass} to 8 classes: \emph{Car}, \emph{Road}, \emph{Sidewalk}, \emph{Crosswalk}, \emph{Curb}, \emph{Person}, \emph{Truck} and \emph{Bus}.
The remaining 2000 panoramas from 40 different cities are collected for omni-supervised learning.
All images have a 360$^{\circ}\times$70$^{\circ}$ FoV at 2048$\times$400.
WildPASS thereby is the largest panoramic street-scene segmentation benchmark, exceeding other evaluation-oriented WildDash (211 annotations)~\cite{zendel2018wilddash}, ISSAFE (313 annotations)~\cite{zhang2020issafe} and PASS (400 annotations)~\cite{yang2020pass} datasets.

\subsection{Experimental Setup}
\label{sec:setup}
Our experiments are built on the Mapillary Vistas~\cite{neuhold2017mapillary} and IDD20K~\cite{varma2019idd} pinhole datasets for a multi-source setting.
These datasets are appealing as they feature diverse viewpoints and unstructured scenes inherent in real-world unconstrained surroundings.
Both datasets contain 18000/2000 and 14027/2036 images for training/validation, respectively. Following~\cite{yang2020omnisupervised}, we use 25 classes from Vistas and the level-3 labels (26 classes) of IDD20K, 15 of which overlap.
We use the 2000 unlabeled panoramas of WildPASS to create omnidirectional annotations with PSPNet50~\cite{zhao2017pyramid} for data distillation.
We test on public PASS (400 annular images) and WildPASS (500 panoramas) datasets and evaluate with the mean Intersection over Union (mIoU) metric.
Facilitating fair comparisons with omnidirectional segmentation frameworks~\cite{yang2020pass,yang2020omnisupervised}, ECANet is implemented using the ERFNet~\cite{romera2018erfnet} backbone pre-trained on ImageNet~\cite{russakovsky2015imagenet}.
We use $N=4$ segments and a strip pooling of $(\widehat{H},\widehat{W})=(2,16)$ in HSA and spatial pooling of $(\widehat{H},\widehat{W})=\{(3,3),(4,4),(6,6)\}$ in PSA.
ECANet is trained under Adam optimization~\cite{kingma2014adam} with a Weight Decay of 2$\times$10$^{-4}$ and a Learning Rate of 1$\times$10$^{-4}$ that decreases exponentially over 200 epochs. We train with a batch-size of 12 and a resolution of 1024$\times$512.
For multi-source training, each iteration comprises a forward and backward pass per dataset with standard cross-entropy loss throughout. 

\begin{table}[t]
\scriptsize
\begin{center}
\begin{tabular}{c|c|c}
\toprule
\multicolumn{2}{c|}{\textbf{Network and Method}}&{\textbf{mIoU}}\\
\midrule
{Vistas-trained}&{Fast-SCNN~\cite{poudel2019fast}}&{28.5\%}\\
{Vistas-trained}&{SegNet~\cite{badrinarayanan2017segnet}}&{25.7\%}\\
{Vistas-trained}&{DRNet (ResNet18)~\cite{yu2017dilated}}&{28.0\%}\\
{Vistas-trained}&{PSPNet (ResNet50)~\cite{zhao2017pyramid}}&{41.4\%}\\
{Vistas-trained}&{DenseASPP (DenseNet121)~\cite{yang2018denseaspp}}&{33.3\%}\\
{Vistas-trained}&{DANet (ResNet50)~\cite{fu2019dual}}&{38.9\%}\\
\midrule
{Style-transferred}&{ENet~\cite{paszke2016enet}}&{31.0\%}\\
{Style-transferred}&{SQNet~\cite{treml2016speeding}}&{27.9\%}\\
{Style-transferred}&{ERFNet~\cite{romera2018erfnet}}&{34.3\%}\\
{Style-transferred}&{LinkNet~\cite{chaurasia2017linknet}}&{30.5\%}\\
{Style-transferred}&{PSPNet (ResNet18)~\cite{zhao2017pyramid}}&{34.8\%}\\
{Style-transferred}&{ICNet~\cite{zhao2018icnet}}&{25.7\%}\\
{Style-transferred}&{ESPNet~\cite{mehta2018espnet}}&{24.7\%}\\
{Style-transferred}&{BiSeNet~\cite{yu2018bisenet}}&{27.7\%}\\
{Style-transferred}&{EDANet~\cite{lo2019efficient}}&{30.5\%}\\
{Style-transferred}&{CGNet~\cite{wu2018cgnet}}&{30.4\%}\\
{Style-transferred}&{SwiftNet~\cite{orvsic2019defense}}&{37.4\%}\\
{Style-transferred}&{SwaftNet~\cite{yang2020ds}}&{38.2\%}\\
\midrule
{Vistas-trained}&{ERF-PSPNet}&{32.2\%}\\
{Style-transferred}&{ERF-PSPNet}&{39.2\%}\\
{PASS~\cite{yang2020pass}}&{ERF-PSPNet}&{58.2\%}\\
{OOSS~\cite{yang2020omnisupervised}}&{ERF-PSPNet}&{47.9\%}\\
\midrule
{Our method}&{ERF-PSPNet}&{\textbf{58.4\%}}\\
{Our method}&{ECANet}&{\textbf{60.2\%}}\\
\bottomrule
\end{tabular}
\end{center}
\vskip-2ex
\setlength{\abovecaptionskip}{0pt}
\setlength{\belowcaptionskip}{0pt}
\caption{Accuracy analysis on the public PASS dataset~\cite{yang2020pass}.}
\label{table_pass}
\vskip-4ex
\end{table}

\subsection{Results on PASS}
\label{sec:pass_results}
As shown in Table~\ref{table_pass}, we first present the results on the public Panoramic Annular Semantic Segmentation (PASS) dataset~\cite{yang2020pass}, which
represents an unseen test-bed to our models. Thereby, the evaluation paradigm expects a trained model that can generalize well in an open, previously unseen domain.
A variety of networks has been benchmarked on the PASS dataset, including some trained on Mapillary Vistas and some style-transferred using CycleGAN~\cite{yang2020pass,yang2020ds}.
Overall, even though the networks under our multi-source omni-supervision have not seen any annular image from the PASS domain, they outperform all previous models by large margins.
In comparison to the PASS method using ring-padding~\cite{payen2018eliminating} for continuous segmentation and the single-pass OOSS method~\cite{yang2020omnisupervised}, our improvement is significant thanks to the diversity introduced in the data distillation.
Moreover, our ECANet capturing omni-range dependencies, boosts mIoU to 60.2\%, setting new state-of-the-art accuracy among 20 mainstream networks.

\begin{table*}[t]
\scriptsize
\begin{center}
\begin{tabular}{c|c|cccccccc|cc}
\toprule
{\textbf{Network}}&{\textbf{mIoU}}&{\textbf{Car}}&{\textbf{Road}}&{\textbf{Sidew.}}&{\textbf{Crossw.}}&{\textbf{Curb}}&{\textbf{Person}}&{\textbf{Truck}}&{\textbf{Bus}}&{\textbf{MACs}}&{\textbf{PARAMs}}\\
\midrule
{SegNet~\cite{badrinarayanan2017segnet}}&{22.7\%}&{57.2\%}&{61.1\%}&{18.2\%}&{4.2\%}&{14.1\%}&{5.8\%}&{9.6\%}&{11.1\%}&{398.3G}&{28.4M}\\
{PSPNet50~\cite{zhao2017pyramid}}&{46.1\%}&{80.0\%}&{74.9\%}&{51.7\%}&{23.9\%}&{31.4\%}&{19.8\%}&{38.9\%}&{48.1\%}&{403.0G}&{53.3M}\\
{DenseASPP~\cite{yang2018denseaspp}}&{33.2\%}&{51.1\%}&{69.1\%}&{38.1\%}&{16.4\%}&{26.3\%}&{8.7\%}&{27.4\%}&{28.4\%}&{78.3G}&{8.3M}\\
{DANet~\cite{fu2019dual}}&{47.2\%}&{74.8\%}&{72.2\%}&{49.9\%}&{28.9\%}&{23.8\%}&{25.4\%}&{51.9\%}&{50.6\%}&{114.1G}&{47.4M}\\
\midrule
{ENet~\cite{paszke2016enet}}&{23.8\%}&{41.9\%}&{58.3\%}&{31.9\%}&{9.2\%}&{17.3\%}&{3.2\%}&{14.8\%}&{13.9\%}&{4.9G}&{0.4M}\\
{Fast-SCNN~\cite{poudel2019fast}}&{24.8\%}&{45.9\%}&{60.0\%}&{31.7\%}&{9.7\%}&{17.1\%}&{6.0\%}&{14.2\%}&{13.8\%}&{1.7G}&{1.1M}\\
{PSPNet18~\cite{zhao2017pyramid}}&{28.2\%}&{58.2\%}&{66.8\%}&{28.4\%}&{13.0\%}&{19.2\%}&{6.2\%}&{18.2\%}&{15.6\%}&{235.0G}&{17.5M}\\
{DRNet22~\cite{yu2017dilated}}&{27.3\%}&{53.0\%}&{66.2\%}&{19.7\%}&{7.1\%}&{16.3\%}&{7.5\%}&{20.5\%}&{28.2\%}&{136.5G}&{15.9M}\\
{ERFNet~\cite{romera2018erfnet}}&{29.8\%}&{64.7\%}&{68.0\%}&{25.9\%}&{6.2\%}&{22.0\%}&{9.0\%}&{19.2\%}&{23.0\%}&{30.3G}&{2.1M}\\
{CGNet~\cite{wu2018cgnet}}&{25.8\%}&{49.7\%}&{60.1\%}&{24.0\%}&{9.9\%}&{15.3\%}&{4.6\%}&{16.1\%}&{26.5\%}&{7.0G}&{0.5M}\\
{SwiftNet~\cite{orvsic2019defense}}&{30.0\%}&{55.7\%}&{64.1\%}&{29.2\%}&{16.2\%}&{22.9\%}&{8.5\%}&{21.1\%}&{22.2\%}&{41.7G}&{11.8M}\\
{SwaftNet~\cite{yang2020ds}}&{35.4\%}&{64.0\%}&{68.5\%}&{37.2\%}&{10.7\%}&{26.7\%}&{13.1\%}&{27.8\%}&{35.6\%}&{41.8G}&{11.9M}\\
{ERF-PSPNet~\cite{yang2020pass}}&{34.0\%}&{66.3\%}&{70.5\%}&{36.5\%}&{6.4\%}&{24.1\%}&{9.4\%}&{26.5\%}&{32.0\%}&{26.6G}&{2.5M}\\
\midrule
{PASS (ERF-PSPNet)~\cite{yang2020pass}}&{64.7\%}&{87.3\%}&{80.0\%}&{61.4\%}&{\textbf{71.1\%}}&{49.9\%}&{\textbf{72.2\%}}&{37.5\%}&{57.9\%}&{91.6G}&{2.5M}\\
{OOSS (ERF-PSPNet)~\cite{yang2020omnisupervised}}&{56.1\%}&{87.2\%}&{79.3\%}&{60.8\%}&{28.0\%}&{38.1\%}&{54.5\%}&{48.8\%}&{52.2\%}&{26.6G}&{2.5M}\\
\midrule
{Our omni-supervised (ERF-PSPNet)}&{66.8\%}&{90.5\%}&{82.7\%}&{65.6\%}&{70.5\%}&{51.5\%}&{58.2\%}&{\textbf{62.0\%}}&{53.1\%}&{26.6G}&{2.5M}\\
{With attention (ECANet)}&{67.7\%}&{90.4\%}&{83.7\%}&{\textbf{68.4\%}}&{67.8\%}&{\textbf{52.1\%}}&{61.4\%}&{54.5\%}&{\textbf{63.3\%}}&{27.8G}&{2.6M}\\
{With fusion (ECANet)}&{\textbf{69.0\%}}&{\textbf{90.6\%}}&{\textbf{85.7\%}}&{68.0\%}&{67.9\%}&{\textbf{52.1\%}}&{66.0\%}&{59.3\%}&{62.3\%}&{27.8G}&{2.6M}\\
\bottomrule
\end{tabular}
\end{center}
\vskip-2ex
\setlength{\abovecaptionskip}{0pt}
\setlength{\belowcaptionskip}{0pt}
\caption{Accuracy analysis on WildPASS.}
\label{table_wildpass}
\vskip-4ex
\end{table*}
\subsection{Results on WildPASS}
\label{sec:results_wildpass}
\noindent
\textbf{Comparison with state-of-the-art.}
With the novel WildPASS dataset, we establish a benchmark for omnidirectional segmentation.
The most accurate and efficiency-oriented architectures are trained on Mapillary Vistas and extensively evaluated on WildPASS.
Table~\ref{table_wildpass} shows their per-class accuracy as well as their computation complexity quantified by Multiply-Accumulate operations (MACs) and number of Parameters.
It can be seen that they generally achieve unsatisfactory results.
For example, DenseASPP is reported to have a mIoU of 65.8\% on Vistas but degrades by more than 30.0\% when taken to our challenging WildPASS.
State-of-the-art panoramic segmentation frameworks PASS~\cite{yang2020pass} and OOSS~\cite{yang2020omnisupervised} improve upon the ERF-PSPNet baseline.
However, PASS needs to separate the panorama into multiple partitions and thereby requires nearly 4 times of MACs, while OOSS is limited without materializing diverse panoramas in data distillation.
Our proposed multi-source omni-supervised approach outperforms both of them while maintaining the high efficiency of ERF-PSPNet.
With concurrent attention, the proposed ECANet surpasses ERF-PSPNet and our multi-space fusion elevates mIoU to 69.0\%, outperforming all previous methods.

\begin{table}[t]
\scriptsize \setlength{\tabcolsep}{2.0pt}
\begin{center}
\begin{tabular}{c|ccc}
\toprule
{\textbf{Network}}&{\textbf{Vistas-trained}}&{\textbf{Omni-supervised}}&{\textbf{Multi-space fused}}\\
\midrule
{ERF-PSPNet~\cite{yang2020pass}}&{34.0\%}&{66.8\%}&{68.1\%}\\
{Fast-SCNN~\cite{poudel2019fast}}&{24.8\%}&{45.4\%}&{45.8\%}\\
{CGNet~\cite{wu2018cgnet}}&{25.8\%}&{48.2\%}&{48.8\%}\\
{DRNet22~\cite{yu2017dilated}}&{27.3\%}&{64.7\%}&{66.0\%}\\
{Our ECANet}&{\textbf{39.7\%}}&{\textbf{67.7\%}}&{\textbf{69.0\%}}\\
\bottomrule
\end{tabular}
\end{center}
\vskip-2ex
\setlength{\abovecaptionskip}{0pt}
\setlength{\belowcaptionskip}{0pt}
\caption{Analysis of our approach for various efficient networks.}
\label{table_various}
\vskip-6ex
\end{table}
\noindent
\textbf{Effectiveness for different networks.}
Furthermore, we experiment with various efficient ConvNets to study how well our approach generalizes.
As shown in Table~\ref{table_various}, the multi-source omni-supervision is advantageous for all efficient networks, as the mIoU scores on WildPASS are improved by significant amounts, even up to 37.4\%.
The multi-space fusion is also consistently effective and the benefit is most pronounced for more accurate networks. These results demonstrate that our novel learning regimen is not strictly tied to a specific architecture and enables reliable predictions for omndirectional imagery across the board.

\begin{table}[t]
\scriptsize \setlength{\tabcolsep}{2.0pt}
\begin{center}
\begin{tabular}{c|c|c}
\toprule
{\textbf{Method}}&{\textbf{on Vistas}}&{\textbf{on WildPASS}}\\
\midrule
{Vistas-trained}&{61.6\%}&{34.0\%}\\
{Multi-source trained}&{63.0\%}&{39.8\%}\\
\midrule
{Single-source fine-tuned}&{20.3\%}&{41.1\%}\\
{Multi-source fine-tuned}&{45.2\%}&{60.7\%}\\
\midrule
{Omni-supervised (Pittsburgh)~\cite{yang2020omnisupervised}}&{62.9\%}&{56.1\%}\\
{Omni-supervised (World-wide panoramas)}&{62.9\%}&{66.8\%}\\
\midrule
{CLAN~\cite{luo2019taking}}&{39.9\%}&{38.3\%}\\
{USSS~\cite{kalluri2019universal}}&{49.5\%}&{38.7\%}\\
{PASS~\cite{yang2020pass}}&{-}&{64.7\%}\\
{Seamless~\cite{porzi2019seamless}}&{-}&{34.0\%}\\
\midrule
{Our method (ERF-PSPNet)}&{62.9\%}&{68.1\%}\\
{Our method with attention (ECANet)}&{\textbf{63.5\%}}&{\textbf{69.0\%}}\\
\bottomrule
\end{tabular}
\end{center}
\vskip-2ex
\setlength{\abovecaptionskip}{0pt}
\setlength{\belowcaptionskip}{0pt}
\caption{Analysis of our multi-source learning method.}
\label{table_training}
\vskip-6ex
\end{table}
\subsection{Ablation Study}
\label{sec:ablation}
\noindent
\textbf{Ablation of multi-source omni-supervision.}
We first analyze our multi-source omni-supervision training method with ERF-PSPNet~\cite{yang2020pass} on Vistas validation set and our WildPASS.
As shown in Table~\ref{table_training}, while the single-source Vistas-learned model is accurate in the training domain (61.6\%), it only yields a mIoU of 34.0\% on WildPASS.
The IDD20K-trained model does not contain all the classes and thus no fair evaluation is possible, yet, the multi-source joint-training boosts the score to 39.8\%.
Fine-tuning on panoramas with generated annotations in both a single- or multi-source fashion improves the results on WildPASS. Omni-supervised learning by using the Pittsburgh dataset~\cite{yang2020omnisupervised} is limited by less diverse panoramic images.
In comparison, omni-supervision on world-wide panoramas largely improves the mIoU to 66.8\%, demonstrating the importance of diverse panoramic images in the data distillation process.
We further compare to the state-of-the-art domain adaptation method CLAN~\cite{luo2019taking} by adapting from pinhole to panoramic imagery and the multi-source semi-supervised method USSS~\cite{kalluri2019universal}. Moreover, PASS~\cite{yang2020pass} for panoramic- and Seamless~\cite{porzi2019seamless} for panoptic segmentation enable a smooth holistic understanding.
However, both of the above frameworks remain sub-optimal in contrast to our method with ERF-PSPNet (68.1\%) and ECANet (69.0\%) which reaches high performances via multi-source omni-supervision and multi-space fusion.

\begin{figure*}[t]
\setlength{\abovecaptionskip}{0pt}
\setlength{\belowcaptionskip}{0pt}
\centering
\includegraphics[width=1.0\textwidth]{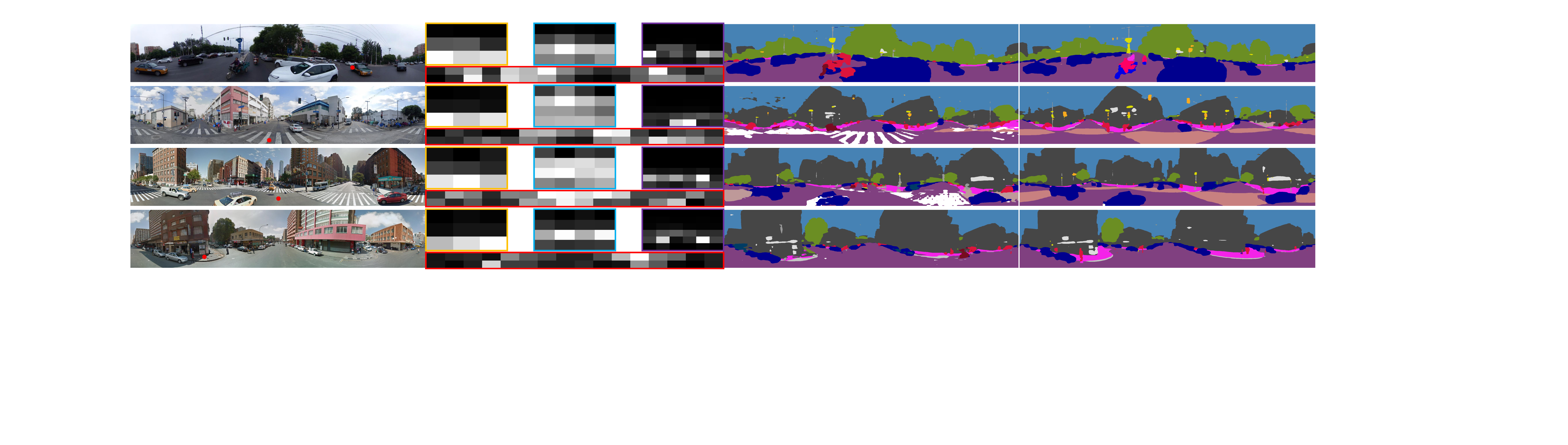}
\captionsetup[subfloat]{captionskip=-10pt}
\subfloat[\label{figurelabel_qualitative_a}Panorama]{\hspace{0.25\textwidth}}
\subfloat[\label{figurelabel_qualitative_b}Sub-attention maps]{\hspace{0.25\textwidth}}
\subfloat[\label{figurelabel_qualitative_c}Baseline]{\hspace{0.25\textwidth}}
\subfloat[\label{figurelabel_qualitative_d}Ours]{\hspace{0.25\textwidth}}
\vskip-1ex
\caption{Qualitative examples of omnidirectional semantic segmentation: (a) Input panorama. (b) Sub-attention maps for the red point in (a), generated with our omni-supervised ECANet including the \RED{\textbf{2$\times$16}} map from the HSA- and the \{\ORANGE{\textbf{3$\times$3}}, \BLUE{\textbf{4$\times$4}}, \PURPLE{\textbf{6$\times$6}}\} maps from the PSA module. The brighter, the higher correlation to the region (maps normalized for visualization). For instance, the red point on the left sidewalk in the last row is correlated to the right sidewalk region according to the 2$\times$16 map. (c) Baseline (ERF-PSPNet~\cite{yang2020pass}). (d) Ours.}
\label{figurelabel_qualitative}
\vskip-2ex
\end{figure*}

\begin{table}[t]
\scriptsize
\begin{center}
\begin{tabular}{c|c|ccc}
\toprule
{\textbf{Network}}&{\textbf{mIoU}}&{\textbf{Memory}}&{\textbf{MACs}}&{\textbf{PARAMs}}\\
\midrule
{ERF-PSPNet~\cite{yang2020pass}}&{34.0\%}&{4.1M}&{4.6G}&{17K}\\
\midrule
{ERFNet~\cite{romera2018erfnet}}&{29.8\%}&{224.0M}&{5.0G}&{189K}\\
{Non-local~\cite{wang2018non}}&{36.6\%}&{261.3M}&{2.6G}&{38K}\\
{Dual attention~\cite{fu2019dual}}&{36.7\%}&{265.3M}&{2.9G}&{75K}\\
{Criss-cross attention~\cite{huang2019ccnet}}&{38.3\%}&{14.5M}&{0.7G}&{79K}\\
{Axial attention~\cite{wang2020axial}}&{39.0\%}&{22.5M}&{0.9G}&{104K}\\
\midrule
{Our ECANet (HSA)}&{39.4\%}&{5.3M}&{0.3G}&{43K}\\
{Our ECANet (PSA)}&{37.5\%}&{14.7M}&{0.9G}&{116K}\\
{ECANet (HSA+PSA)}&{\textbf{39.7\%}}&{20.0M}&{1.3G}&{159K}\\
\bottomrule
\end{tabular}
\end{center}
\vskip-2ex
\setlength{\abovecaptionskip}{0pt}
\setlength{\belowcaptionskip}{0pt}
\caption{Ablation of efficient concurrent attention. Computation metrics correspond to the context aggregation of single-source trained networks after ERFNet backbone (feature res. at 128$\times$64).}
\label{table_ca}
\vskip-4ex
\end{table}

\noindent
\textbf{Ablation of attention and fusion.}
Table~\ref{table_ca} shows the mIoU, memory and computation requirement of our ECANet compared to its dissected versions using only HSA or PSA as well as non-local techniques from literature.
It can be seen that our ECANet, while strengthening the horizontal context aggregation, is much more memory-efficient as compared to the non-local and dual attention modules.
Moreover, ECANet clearly outperforms them as well as efficient axial- and criss-cross attention in omnidirectional segmentation.
This demonstrates ECANet, by capturing essential omni-range contextual priors, is inherently robust in panoramic image segmentation.
In the multi-source omni-supervised setting, we further compare different attention and context aggregation modules (Table~\ref{table_ca_os}).
We experiment with $N=\{2,4,8\}$ segments and verify that a 4-segment ECANet is most accurate.
Moreover, our multi-space fusion method exceeds calibration-bypassed fusion methods that simply take the class with maximum probability from both spaces or from the space with minimum variance.
\begin{table}[t]
\scriptsize
\begin{center}
\begin{tabular}{lc?lc}
\toprule
{\textbf{Network/Method}}&{\textbf{mIoU}}&{\textbf{Network/Module}}&{\textbf{mIoU}}\\
\midrule
{ERF-PSPNet~\cite{yang2020pass}}&{66.8\%}&{Non-local~\cite{wang2018non}}&{62.6\%}\\
\cmidrule{1-2}
{ECANet (HSA, 2 Segs)}&{66.8\%}&{Dual attention~\cite{fu2019dual}}&{65.0\%}\\
{ECANet (HSA, 4 Segs)}&{\textbf{67.4\%}}&{Criss-cross attention~\cite{huang2019ccnet}}&{66.3\%}\\
{ECANet (HSA, 8 Segs)}&{66.7\%}&{Axial attention~\cite{wang2020axial}}&{64.3\%}\\
{ECANet (PSA)}&{67.2\%}&{Height-driven attention~\cite{choi2020cars}}&{64.8\%}\\
{ECANet (HSA+PSA)}&{\textbf{67.7\%}}&{Point-wise attention~\cite{zhao2018psanet}}&{65.4\%}\\
\cmidrule{1-2}
{Fusion (Max Prob.)}&{68.8\%}&{ASPP~\cite{chen2017deeplab}}&{65.7\%}\\
{Fusion (Min Var.)}&{68.4\%}&{BFP~\cite{ding2019boundary}}&{65.6\%}\\
{Our Fusion}&{\textbf{69.0\%}}&{Strip pooling~\cite{hou2020strip}}&{63.7\%}\\
\bottomrule
\end{tabular}
\end{center}
\vskip-2ex
\setlength{\abovecaptionskip}{0pt}
\setlength{\belowcaptionskip}{0pt}
\caption{Accuracy analysis of efficient concurrent attention and fusion methods in the multi-source omni-supervised setting, and comparison against state-of-the-art context aggregation modules.}
\label{table_ca_os}
\vskip-4ex
\end{table}

\subsection{Qualitative Analysis}
\label{sec:qualitative}
Fig.~\ref{figurelabel_qualitative} displays qualitative omnidirectional segmentation examples.
For each panorama, we select a point and show their corresponding attention maps yielded from our ECANet.
It can be observed that ECANet achieves more reliable segmentation than the pinhole-trained baseline, \eg, ours enables complete segmentation of the crosswalk rather than wrongly classifies into general road markings.
The 3$\times$3 attention map (orange) learned to highlight regions within the same height, which shows that it is meaningful to emphasize horizontal dependencies, forming a more discriminative feature representation.
The 4$\times$4 (blue) and 6$\times$6 (purple) maps handle more detailed dependencies.
On the other hand, the 2$\times$16 (red) attention map captures semantically related regions in the horizontal direction stretching across 360$^\circ$.
For example, for the marked points in the first and last row, the spotlighted regions correspond to the strongly correlated homogeneous semantics, \ie, other cars and sidewalks, whose features are passed back, significantly improving the consistency of the segmentation.

\section{Conclusion}
We look into omnidirectional image segmentation from the context-aware perspective.
To bridge the gap in terms of FoV and structural distribution, our proposed ECANet captures inherent omni-range dependencies that stretch across 360$^{\circ}$, whose generalization is optimized via multi-source omni-supervision and multi-space fusion.
To foster progress in panoramic semantic perception, we establish and extensively evaluate models on WildPASS, a benchmark that captures diverse scenes from all around the world.
Comprehensive experimental results show the proposed methods significantly elevate state-of-the-art accuracy for high-efficiency ConvNets on PASS and WildPASS.

\clearpage


{\small
\bibliographystyle{ieee_fullname}
\bibliography{egbib}
}

\end{document}